\documentclass{article}

\usepackage{arxiv}

\usepackage[utf8]{inputenc} 
\usepackage[T1]{fontenc}    
\usepackage{hyperref}       
\usepackage{url}            
\usepackage{booktabs}       
\usepackage{amsfonts}       
\usepackage{nicefrac}       
\usepackage{microtype}      
\usepackage{lipsum}
\usepackage{graphicx}
\graphicspath{ {./images/} }

\usepackage{amssymb}
\usepackage{multirow}

\title{BERT Based Clinical Knowledge Extraction for Biomedical Knowledge Graph Construction and Analysis}

\author{
Ayoub Harnoune \\
  School of Information Sciences\\
  Meridian Team\\
  LyRICA Laboratory \\
   \And
 Maryem Rhanoui \\
  School of Information Sciences\\
  Meridian Team\\
  LyRICA Laboratory \\
   \And
 Mounia Mikram \\
  School of Information Sciences\\
  Meridian Team\\
  LyRICA Laboratory \\
  \And
 Siham Yousfi \\
  School of Information Sciences\\
  Meridian Team\\
  LyRICA Laboratory \\
  \And
 Zineb Elkaimbillah \\
  Rabat IT Center, ENSIAS\\
  Mohammed V University in Rabat\\
  IMS Team, ADMIR Laboratory\\
  \And
 Bouchra El Asri \\
  Rabat IT Center, ENSIAS\\
  Mohammed V University in Rabat\\
  IMS Team, ADMIR Laboratory\\
}

\begin{document}
\maketitle
\begin{abstract}
\noindent \textit{Background : }Knowledge is evolving over time, often as a result of new discoveries or changes in the adopted methods of reasoning. Also, new facts or evidence may become available, leading to new understandings of complex phenomena. This is particularly true in the biomedical field, where scientists and physicians are constantly striving to find new methods of diagnosis, treatment and eventually cure. Knowledge Graphs (KGs) offer a real way of organizing and retrieving the massive and growing amount of biomedical knowledge.

\noindent \textit{Objective :} We propose an end-to-end approach for knowledge extraction and analysis from biomedical clinical notes using the Bidirectional Encoder Representations from Transformers (BERT) model and Conditional Random Field (CRF) layer.

\noindent \textit{Methods :} The approach is based on knowledge graphs, which can effectively process abstract biomedical concepts such as relationships and interactions between medical entities. Besides offering an intuitive way to visualize these concepts, KGs can solve more complex knowledge retrieval problems by simplifying them into simpler representations or by transforming the problems into representations from different perspectives. We created a biomedical Knowledge Graph using using Natural Language Processing models for named entity recognition and relation extraction. The generated biomedical knowledge graphs (KGs) are then used for question answering.

\noindent \textit{Results :}  The proposed framework can successfully extract relevant structured information with high accuracy (90.7\% for Named-entity recognition (NER), 88\% for relation extraction (RE)), according to experimental findings based on real-world 505 patient biomedical unstructured clinical notes.

\noindent \textit{Conclusions :}
In this paper, we propose a novel  end-to-end system for the construction of a biomedical knowledge graph from clinical textual using a variation of BERT models.
\end{abstract}


\section{Introduction}
In the healthcare industry, AI has provided healthcare institutions and practitioners with tools that can be used to reduce the pressure on their workloads and redefine their workflows \cite{davenport2019potential}. 

The challenge raised by the exploitation of medical textual data is that the information on patients is diverse and massive \cite{rencis2019natural}, making some of the tasks performed by health professionals repetitive, long and tedious such as information retrieval. In fact, hospitals are generating massive amounts of data from which new knowledge can be generated. This tremendous increase in the amount of data available now makes it almost impossible for physicians to properly understand and extract new knowledge without assistance. 

The use of efficient computational methods to create knowledge representations is an appropriate alternative to support information retrieval in the medical domain. For example, such representations could provide a way to correlate a concept from a specific work with other concepts from the same study, as well as with concepts from similar investigations. By observing these relationships, physicians and researchers might be able to formulate new hypotheses or draw new conclusions, thus advancing the state of the art in a research area.

This need has led to the search for other more innovative and rapid solutions such as knowledge graph-based information retrieval. Textual, and unstructured data, covers non-trivial latent knowledge \cite{fan2012automatic} that provides significant value for analysis \cite{mooney2005mining}. Retrieving information from such unstructured data is acknowledged as one of the most untapped opportunities in data science.

Knowledge Discovery is the process of transforming low-level raw data into high-level insightful knowledge \cite{maimon2005data}. To represent this complex knowledge, using a graph-based abstraction offers numerous benefits when compared with traditional representations, as relational model or NoSQL alternatives. Graphs provide a concise and intuitive abstraction for a variety of domains, where edges capture the relations between the entities \cite{hogan2021knowledge}.

In this context, knowledge graphs, inspired by human intelligence and problem solving, is a powerful structured representation of facts, consisting of entities, relationships, and semantic descriptions \cite{ji2021survey}. It is particularly useful in the health and biomedical field \cite{shi2017semantic, wang2020recent}.

Transformers \cite{vaswani2017attention} are a type of neural network architecture that has gained popularity in recent years due to their high performances in various domains as Natural Language Processing (NLP) \cite{wolf2020transformers} and Computer Vision \cite{khan2021transformers}. Specifically, an increasing attention has been brought to the Bidirectional Encoder Representations from Transformers (BERT) \cite{devlin2018bert} pre-trained model that can be easily fine-tuned and achieves state-of-the-art performances for a wide range of NLP tasks.

In this paper, we propose an end-to-end system for the construction of a biomedical knowledge graph from clinical textual, unstructured, and thus difficult to process data, this, using a variation of BERT models. The results of the construction and analysis of this knowledge graph shall provide a basis for medical decisions. This would not only assist healthcare professionals to review the individual entities, but also all the relationships between them. And would also help physicians easily find the relationships between a drug and clinical notes, so that these drugs can be closely monitored.

Moreover, compared to the existing works, most of them focus on a specific part in the life cycle of the construction of a knowledge graph, and thus they are not combining these elements for the global construction of the graph and at times they do not include the analysis part to illustrate the interest of the construction. Thus, in this paper, we cover the graph life cycle starting with the construction part and ending with the graph analysis part.

The main contributions are summarized as follows:

\begin{itemize}
\item we built an end-to-end system that automatically structures clinical data into a knowledge graph format that would help physicians and patients quickly find the information they need. 
\item concretely, we built a named entity recognition model that would recognize entities such as drug, strength, duration, frequency, adverse drug reactions, reason for taking the drug, route of administration and form. In addition, we built a model that would also recognize the relationship between the drug and any other named entity. 
\end{itemize}

The remainder of this paper is organized as follows. In Section two, we present the technical background, then we summarize in section three the related works. Section four presents our proposed approach and finally the last section discusses the results.

\section{Background}
\label{sec:back}

\subsection{Knowledge Graph and Applications}

A knowledge graph (KG) is a representation of knowledge related to a domain in a machine-readable form. It is a directed labeled graph in which the labels have well-defined meanings. A directed labeled graph is composed of nodes, edges and labels.

The knowledge graph represents a collection of related descriptions of entities - objects and events in the real world where :
\begin{itemize}
\item Descriptions have formal semantics enabling people and computers to process them efficiently and unambiguously.
\item Descriptions of entities contribute to each other, forming a network, where each entity represents a part of the description of entities related to it, and provides a context.
\end{itemize}
Knowledge graphs are used for a wide range of applications \cite{zou2020survey} from spatial, journalism, biomedical \cite{rotmensch2017learning, xu2020building, nicholson2020constructing, moon2021learning, zhang2021drug}  to education \cite{chen2018knowedu, de2020education, elkaimbillah2021comparative}, recommender systems \cite{wang2019kgat, zhang2019deep, cao2019unifying, yang2020hagerec} and pharmaceuticals \cite{abdelaziz2017large, karim2019drug, zeng2020network}.

However, among biomedical applications, knowledge graphs for clinical analysis and patient care is still at an early stage despite its considerable interest. These graphs can be constructed from clinical notes or electronic health records, where nodes represent patients, drugs and diseases while edges represent relationships such as a patient being prescribed a treatment or a patient being diagnosed with a disease \cite{nicholson2020constructing}.

\subsection{Knowledge Graph Construction}

\subsubsection{Named Entity Recognition}

Named Entity Recognition (NER) (also known as Named Entity Identification, Entity Clipping, and Entity Extraction) is a subtask of information retrieval that aims to locate and classify named entities mentioned in unstructured text into predefined categories such as people's names, organizations, places, medical codes, time expressions, quantities, monetary values, percentages, etc.

The three main classes are: the entity class (such as person name, place name, and institution name), the time class (such as date and time), and the number class (for example, currency and percentage). These classes can be extended to fit specific application areas.

\subsubsection{Coreference Resolution}

Coreference resolution (CR) is a challenging task in natural language processing (NLP). It aims at grouping expressions that refer to the same real-world entity in order to obtain a less ambiguous text. It is useful in tasks such as text comprehension, question answering and summarization.

\subsubsection{Relation Extraction}

Relation extraction (RE) is the extraction of semantic relationships in a text. Extracted relationships typically occur between two or more entities of some type (e.g., person, organization, place) and are classified into a number of semantic categories (e.g., married to, employed by, lives in).

\subsection{Transformers}

Transformers \cite{wolf2019huggingface, wolf2020transformers} are a type of neural network architecture that has gained popularity in recent years. They were developed to solve the problem of sequence transduction, or neural machine translation. This means any task that transforms an input sequence into an output sequence, including speech recognition, text-to-speech transformation, etc.

The basic architecture of transformers consists of a stack of encoders fully connected to a stack of decoders. Each encoder consists of two blocks: a self-attentive component and an anticipatory network. Each decoder consists of three blocks: a self-attention component, an encoder-decoder attention component and an anticipation component.

\begin{figure}[htbp]
\centering
  \includegraphics[width=3in]{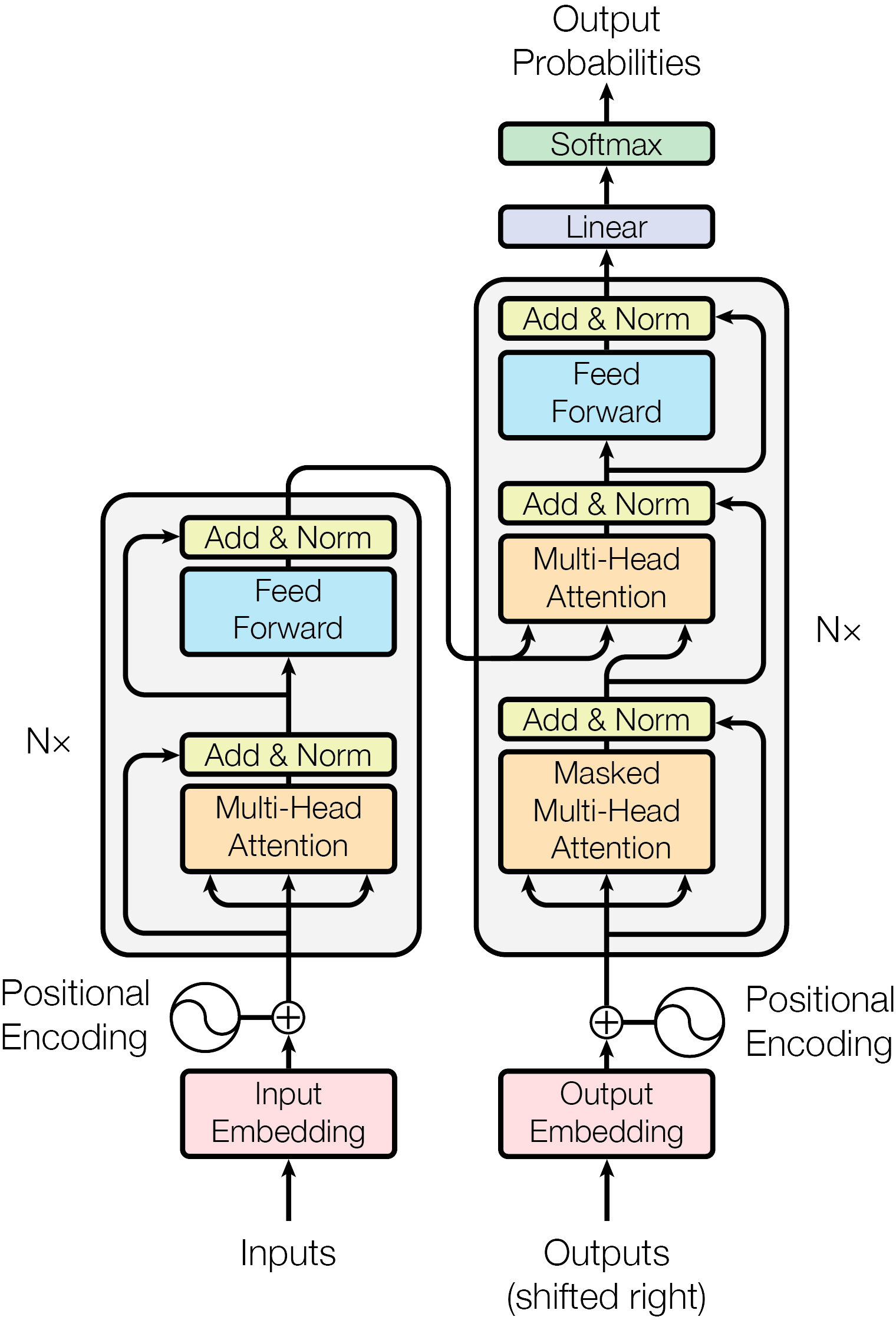}
 \caption{Transformers Architecture \cite{vaswani2017attention}}
  \label{fgr:trans}
\end{figure}

\subsubsection{BERT}

BERT \cite{devlin2018bert} stands for Bidirectional Encoder Representations from Transformers. This model is both :
\begin{itemize}
\item More efficient than its predecessors in terms of learning speed.
\item Is adaptable, based on its initial configuration, to a particular task, such as the task of extracting relations.
\end{itemize}

One of the main reasons for BERT's excellent performance on different NLP tasks is the the possibility of using semi-supervised learning. This translates into the fact that the model is trained to a particular task that allows it to grasp the linguistic patterns. 
After training, the model is equipped with language processing features that can be used to consolidate other models we develop and train. models that we develop and train using supervised learning.

\subsubsection{BERT Variants}

In the following, we will explore several major versions of BERT.

\paragraph{DistilBERT}

DistilBERT \cite{sanh2019distilbert} is a distilled version of BERT, it is smaller, faster, cheaper and lighter.
Distil-BERT has 97\% of the performance of BERT while being trained on half of the parameters of BERT. BERT-base has 110 parameters and BERT-large has 340 parameters, which is difficult to manage. To solve this problem, the distillation technique is used to reduce the size of these large models.
The general architecture of Distil-BERT is the same as that of BERT, except that token embeddings and the pooler are removed, while reducing the number of layers by a factor of 2, which has a significant impact on computational efficiency.

\paragraph{BioBERT}

BioBERT \cite{lee2020biobert} is a variant of BERT that is pretrained on biomedical dataset. In the pretraining, the weights of the regular BERT model were taken and then pre-trained on the medical datasets like (PubMed abstracts and PMC). This domain specific pretrained model can be refined for many tasks such as Named Entity Recognition (NER), Relationship Extraction (RE) and Question Answering System (QA).

\paragraph{BioClinicalBERT}

Clinical notes contain information about patients that goes beyond structured data such as laboratory values and medications. However, clinical notes have been underutilized compared to structured data because they are highly dimensional and sparse.
ClinicalBERT \cite{alsentzer2019publicly} is a variant of BERT that specializes in clinical notes. It highlights high quality relationships between medical concepts as judged by humans.
This variant is pretrained on the Medical Information Mart for Intensive Care III (mimiciii) dataset. mimic-iii consists of the electronic medical records of 58,976 unique hospital admissions of 38,597 patients in the intensive care unit between 2001 and 2012. There are 2,083,180 de-identified notes associated with the admissions

\paragraph{Bio-Discharge-Summary}

Bio-Discharge-Summary \cite{alsentzer2019publicly} is a biomedical-specific variant of BERT, which was initialized from BioBERT and trained on the clinical notes from the MIMIC hospital. It was then refined for the five tasks with minimal architectural modifications. As with BERT, the large (24 layers, 16 attention heads, output integration size of 1024) and basic (12 layers, 12 attention heads, output integration size of 768) versions of this model were pre-trained and refined.

\section{Related Works}

Several works applied Deep Learning to automate the construction of a knowledge graph, some focused on the named entity recognition task, others chose to address relationship extraction, and many others dealt with the coreference resolution problem. The related works are summarized in the Table \ref{tab:rw}.

Therefore, this section is organized along three axes:

\subsection{Named Entity Recognition}

Li et al. \cite{li2017neural} explore a joint neural model for extracting biomedical entities and their relationships. Their model uses the advantages of several state-of-the-art neural models for entity recognition or relation classification in text mining and NLP. Their model was evaluated on two tasks, namely, extracting drug-related adverse events between the entities "drug" and "disease", and extracting residency relationships between the entities "bacteria" and "location".

Copara et al. \cite{copara2020contextualized} studied contextualized language models for French NER in clinical texts in the context of the Text Mining Challenge (TMC) knowledge extraction task. This task is divided into two subtasks, which aim to identify the entities anatomy, dose, examination, mode, time, pathology, sosy, substance, treatment and value in clinical cases. Since each language has its own particularities, their hypothesis is that it is interesting to design a specific linguistic model for French clinical corpora. Thus, they explore a model based on CamemBERT pre-trained on a French biomedical corpus and refined on the data of the DEFT information extraction task.

Jiang et al. \cite{jiang2019bert} applied a BERT-BiLSTM-CRF model to the named entity recognition of Chinese electronic medical records. This model improves the semantic representation of words by using the BERT pre-trained language model, then they combine a BiLSTM network with a CRF layer, and the word vector is used as input for model training. The BERT model can improve the semantic representation of sentences, the BiLSTM network can effectively solve the problem of previous methods that rely on domain knowledge and feature engineering, and the CRF can focus on context annotation information compared to other models.

Lample et al. \cite{lample2016neural} introduced two new neural architectures, one based on bidirectional LSTMs and conditional random fields, and one that constructs and labels segments using a transition-based approach inspired by shift-reduce parsers. Their models rely on two sources of word information: character-based word representations learned from the supervised corpus, and unsupervised word representations learned from unannotated corpora. Their models achieve state-of-the-art NER performance in four languages without resorting to language-specific knowledge or resources.

Akbik et al. \cite{akbik2018contextual} proposed to exploit the internal states of a trained character language model to produce a new type of word embedding that they call contextual string embedding. Their proposed embeddings have the following distinct properties: (a) they are trained without an explicit notion of words and thus fundamentally model words as sequences of characters, and (b) they are contextualized by the surrounding text, meaning that the same word will have different embeddings depending on its contextual use. They performed a comparative evaluation against previous embeddings and found that their embeddings are very useful for downstream tasks in particular, on named entity recognition (NER) in English and German.

Ma et al. \cite{ma2016end} proposed a novel neutral network architecture that automatically benefits from word- and character-level representations, using a combination of bidirectional LSTM, CNN, and CRF. Their system is truly end-to-end, requiring no feature engineering or data preprocessing, making it applicable to a wide range of sequence labeling tasks.

\begin{table}[htht]
\footnotesize
\caption{Related Works}
\label{tab:rw}
\begin{tabular}{lllllll}
\hline
\multicolumn{7}{c}{\textbf{NER}} \\ \hline
\multirow{2}{*}{\textbf{Ref}} &
  \multirow{2}{*}{\textbf{Corpus}} &
  \multicolumn{2}{l}{\textbf{Input Representation}} &
  \multirow{2}{*}{\textbf{\begin{tabular}[c]{@{}l@{}}Context \\ Encoder\end{tabular}}} &
  \multirow{2}{*}{\textbf{Tag Encoder}} &
  \multirow{2}{*}{\textbf{\begin{tabular}[c]{@{}l@{}}Perf\\ F1 score\end{tabular}}} \\ \cline{3-4}
 &
   &
  \textbf{Character} &
  \textbf{Word} &
   &
   &
   \\ \hline
\cite{li2017neural} &
  UMLS &
  CNN &
  Glove &
  BiLSTM &
  Softmax &
  71\% \\ \hline
\cite{copara2020contextualized}   &
  DEFT 2020 &
  X &
  \begin{tabular}[c]{@{}l@{}}Wordpiece \\ (word\\ +subword)\end{tabular} &
  \begin{tabular}[c]{@{}l@{}}Bert multilingual cased\\ Camembert-bio large \\ Camembert large\end{tabular} &
  Softmax &
  \begin{tabular}[c]{@{}l@{}}58\%\\ 65\%\\ 67\%\end{tabular} \\ \hline
\cite{jiang2019bert}  &
  CCKS 2017 &
  LSTM &
  Glove &
  Bert+BiLSTM &
  CRF &
  83\% \\ \hline
\cite{lample2016neural} &
  CoNLL-2003 &
  X &
  Glove &
  BiLSTM &
  CRF &
  91.47\% \\ \hline
\cite{akbik2018contextual} &
  CoNLL-2003 &
  X &
  \begin{tabular}[c]{@{}l@{}}Flair \\ embedding\end{tabular} &
  BiLSTM &
  CRF &
  93.89\% \\ \hline
\cite{ma2016end} &
  CoNLL-2003 &
  CNN &
  Glove &
  BiLSTM-CNN &
  CRF &
  91.87\% \\ \hline
\multicolumn{7}{c}{\textbf{RE}} \\ \hline
\multirow{2}{*}{\textbf{Ref}} &
  \multirow{2}{*}{\textbf{Corpus}} &
  \multicolumn{2}{l}{\textbf{Input Representation}} &
  \multirow{2}{*}{\textbf{\begin{tabular}[c]{@{}l@{}}Context \\ Encoder\end{tabular}}} &
  \multirow{2}{*}{\textbf{\begin{tabular}[c]{@{}l@{}}Classification\\ Layer\end{tabular}}} &
  \multirow{2}{*}{\textbf{\begin{tabular}[c]{@{}l@{}}Perf\\ F1 score\end{tabular}}} \\ \cline{3-4}
 &
   &
  \textbf{Character} &
  \textbf{Word} &
   &
   &
   \\ \hline
\cite{wu2019enriching} &
  SemEval 2010 &
  X &
  SentencePiece &
  BERT &
  Softmax &
  89.25\% \\ \hline
\cite{wang2016relation} &
  SemEval 2010 &
  X &
  Word2vec &
  Word2vec &
  \begin{tabular}[c]{@{}l@{}}Att-Pooling-CNN\end{tabular} &
  88\% \\ \hline
\cite{lee2019semantic} &
  SemEval 2010 &
  X &
  \begin{tabular}[c]{@{}l@{}}Word representation\\ + self attention\\ + BiLSTM\end{tabular} &
  \begin{tabular}[c]{@{}l@{}}Word representation\\ + self attention\\ + BiLSTM\end{tabular} &
  \begin{tabular}[c]{@{}l@{}}Entity aware \\ attention \\ (softmax)\end{tabular} &
  85.2\% \\ \hline
  
\cite{xiao2016semantic} &
  SemEval 2010 &
  X &
  \begin{tabular}[c]{@{}l@{}}BiLSTM + position\\ features\end{tabular} &
  \begin{tabular}[c]{@{}l@{}}BiLSTM + Neural \\attention\end{tabular} &
  Softmax &
  84.3\% \\ \hline
  
\cite{eberts2021end} &
  DocRED &
  CNN &
  Wordpiece &
  BERT &
  Softmax &
  92\% \\ \hline
\multicolumn{7}{c}{\textbf{CR}} \\ \hline
\multirow{2}{*}{\textbf{Ref}} &
  \multirow{2}{*}{\textbf{Corpus}} &
  \multicolumn{2}{l}{\textbf{Input Representation}} &
  \multirow{2}{*}{\textbf{Model}} &
  \multirow{2}{*}{\textbf{}} &
  \multirow{2}{*}{\textbf{\begin{tabular}[c]{@{}l@{}}Perf\\ F1 score\end{tabular}}} \\ \cline{3-4}
 &
   &
  \textbf{Character} &
  \textbf{Word} &
   &
   &
   \\ \hline
\cite{joshi2019bert} &
  CoNLL 2012 &
  X &
  Wordpiece &
  BERT &
   &
  76.9\% \\ \hline
\cite{xu2020revealing} &
  CoNLL 2012 &
  X &
  Wordpiece &
  BERT+HOI &
   &
  80.2\% \\ \hline
\cite{joshi2020spanbert} &
  CoNLL 2012 &
  X &
  SentencePiece &
  SpanBERT &
   &
  76.6\% \\ \hline
\end{tabular}%
\end{table}

\subsection{Relation Extraction}

Wu et al. \cite{wu2019enriching} proposed a model that exploits both the pretrained BERT language model and incorporates information from the target entities to tackle the relationship classification task. They locate the target entities and transfer the information through the pretrained architecture and incorporate the corresponding encoding of both entities. 

Wang et al. \cite{wang2016relation} proposed a convolutional neural network architecture for the relation extraction task, which relies on two levels of attention to better discern patterns in heterogeneous contexts. This architecture helps end-to-end learning from task-specific labeled data without the need for external knowledge such as explicit dependency structures.

Lee et al. \cite{lee2019semantic} proposed a new end-to-end recurrent neural model that incorporates an entity-sensitive attention mechanism with a latent entity typing (LET) method. Not only does their model effectively use entities and their latent types as features, but it is also easier to interpret by visualizing the attention mechanisms applied to their model and the results of the LET method.

Xiao et al. \cite{xiao2016semantic} introduced a hierarchical recurrent neural network capable of extracting information from raw sentences for relationship classification. their model has several distinguishing features: (1) Each sentence is divided into three contextual subsequences according to two annotated nominals, which helps the model encode each contextual subsequence independently to selectively focus on important contextual information; (2) The hierarchical model consists of two recurrent neural networks (RNNs): the first one learns the contextual representations of the three contextual subsequences, respectively, and the second one computes the semantic composition of these three representations and produces a sentence representation for the relationship classification of the two nominals. (3) The attention mechanism is adopted in both RNNs to encourage the model to focus on important information when learning sentence representations.

Eberts et al. \cite{eberts2021end} present a joint model for extracting entity-level relations from documents. Unlike other approaches that focus on intra-sentence local mention pairs and thus require mention-level annotations, their model operates at the entity level. For this purpose, they adopt a multi-task approach that relies on coreference resolution and gathers relevant signals via multi-instance learning with multi-level representations combining global information about entities and local mentions.

\subsection{Coreference Resolution}

Joshi et al. \cite{joshi2019bert} applied BERT for coreference resolution, they replaced the entire LSTM-based encoder (with ELMo and GloVe embeddings as input) with the BERT transformer. They addressed the first and last word chunks (concatenated with the assisted version of all word chunks in the interval) as interval representations. The documents are divided into segments of maximum length. A qualitative analysis of the model predictions indicates that, compared to ELMo and BERT-base, BERT-large performs particularly well in distinguishing related but distinct entities.

Xu et al. \cite{xu2020revealing} analyzed the impact of higher order inference (HOI) on the coreference resolution task. HOI has been adapted by almost all recent models of coreference resolution without its actual effectiveness on representation learning having been studied. To perform a comprehensive analysis, They implemented an end-to-end coreference system as well as four HOI approaches, assisted antecedent, feature equalization, extent clustering, and clustering fusion.

Joshi et al. \cite{joshi2020spanbert} presented SpanBERT which is a pre-training method designed to better represent and predict text spans. Their approach extends BERT by (1) masking contiguous random extents, rather than random tokens, and (2) training extent boundary representations to predict the entire content of the masked extent. SpanBERT consistently outperforms the original BERT, with substantial gains in span selection tasks such as question answering and coreference resolution. 

\subsection{Synthesis and Discussion}

The objective of this paper is the construction and analysis of a biomedical knowledge graph from textual data represented by clinical notes, to do this we compare the important versions of BERT in the biomedical domain and we extend the work \cite{copara2020contextualized} who tested and applied the variants of BERT in the medical domain for clinical cases written in French for the task of named entity recognition. Therefore, we combine these BERT variants with other layers namely the CRF layer for the named entity detection task in order to maximize the performance.

\section{Proposed Model}

\begin{figure}[htbp]
  \centering
\centerline{\includegraphics[width=5in]{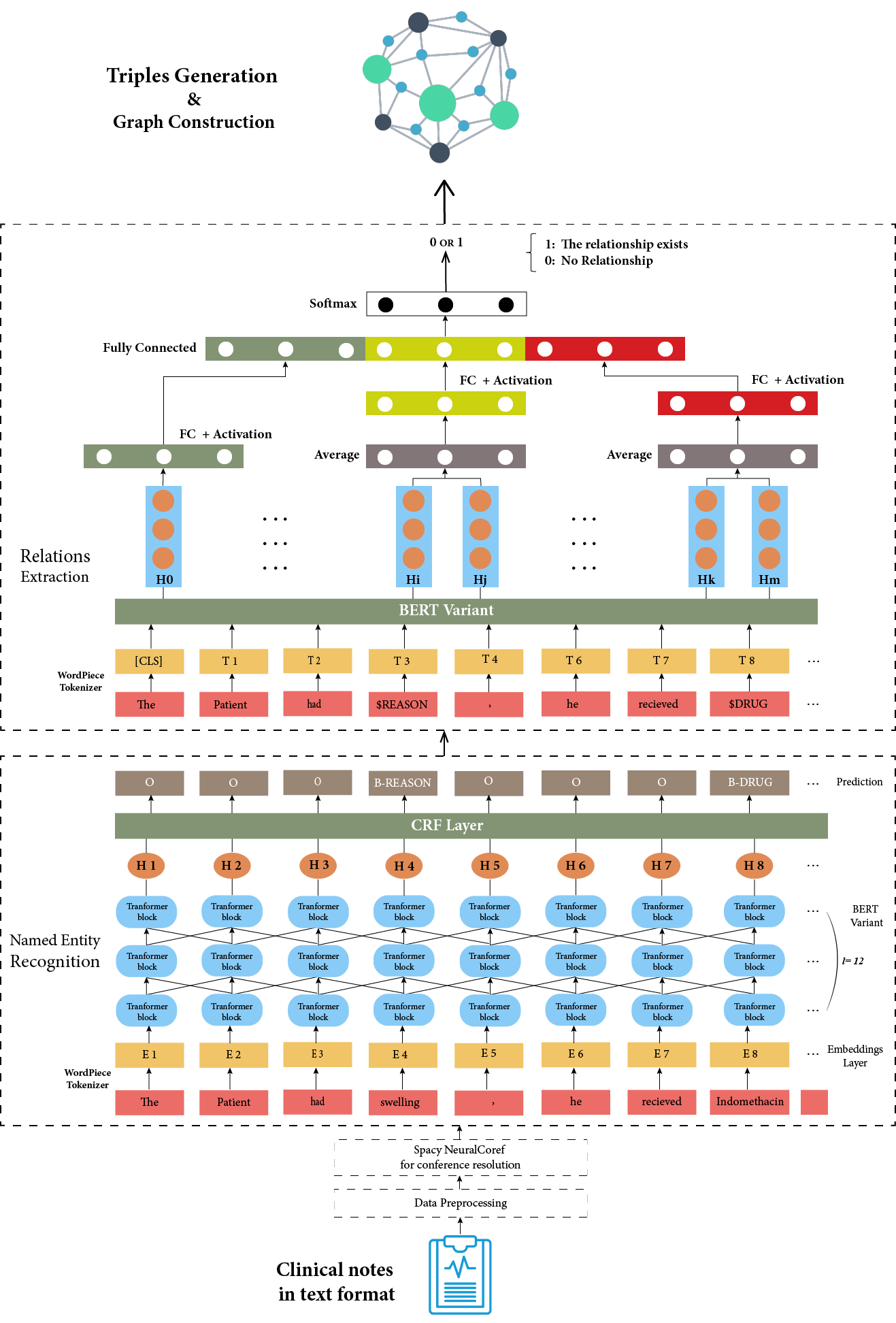}}
  \caption{Proposed Architecture}
\label{fig:model}
 \end{figure}

We propose a generic approach for the construction of a knowledge graph from any clinical textual data written in English language in order to analyze it.

Based on the literature review, BERT outperforms other CNN or RNN models in the three life cycle tasks of building a knowledge graph. Therefore, we test and compare different variants of BERT for the biomedical domain.

\subsection{General Architecture}

The architecture is organized in several phases (fig \ref{fig:model}):
 
\begin{itemize}
\item Preprocessing
\begin{itemize}
\item Preparing the clinical note data for the coreference resolution task.
\end{itemize}
\item Coreference Resolution
\begin{itemize}
\item Solve the coreference using the NeuralCoref model.
\end{itemize}
\item Named Entity Recognition
\begin{itemize}
\item The outputs of the coreference resolution system, pass to the named entity extraction model to detect the entities and its classes.
\end{itemize}
\item Relation Extraction
\begin{itemize}
\item Then, predicting the existence and types of relationships between the named entities predicted by the named entity extraction model by the relationship extraction model.
\end{itemize}
\item Analysis
\begin{itemize}
\item And finally, storing the prediction results in the Neo4j database for analysis.
\end{itemize}
\end{itemize}

\subsection{Preprocessing}

\subsubsection{BERT Embedding}

Like most deep learning models, BERT passes each input token (the words in the input text) through a token embedding layer so that each token is transformed into a vector representation. Unlike other deep learning models, BERT has additional embedding layers in the form of segment embedding and position embedding.

So a tokenized input sequence of length there are three distinct representations, namely:
\begin{itemize}
\item Token Embeddings with the form (1, n, 768) which are just vector representations of words.
\item Segment Embeddings with the form (1, n, 768) which are vector representations to help BERT distinguish matched input sequences.
\item Position Embeddings with the form (1, n, 768) to let BERT know that the inputs provided to it have a temporal property.
\end{itemize}

These representations are summed by elements to produce a single representation of form (1, n, 768). This is the input representation that is passed to BERT's Encoder layer.

\subsection{Knowledge Extraction}

\subsubsection{Named Entity Recognition}

Clinical notes are usually long, and it is not recommended to have such large input sizes for machine learning models, especially for models like BERT that have an input size restriction of 512 tokens. Therefore, a function was implemented to split the clinical note records based on a maximum sequence length parameter. The function attempts to include a maximum number of tokens, keeping as much context as possible for each token. Split points are decided based on the following criteria:

\begin{itemize}
\item Includes as many paragraphs as possible within the maximum number of tokens, and splits at the end of the last paragraph found.
\item If the function does not find a single complete paragraph, it splits on the last line (within the token limit) that marks the end of a sentence.
\item Otherwise, the function includes as many tokens as specified by the token limit, and then splits on the next index.
\end{itemize}

The data is tokenized using the BERT base tokenizer. Each sequence of labels or tokens in the data was represented using the IOB2 (Inside, Outside, Beginning) tagging scheme for BERT variant models.

To obtain optimal results, we apply and compare the variants of BERT to select the model that gives the best performance in the NER and RE tasks.

In this context, we use BERT variants in the medical domain for context encoding namely: BioBERT, BioClinicalBERT, BioDischargeSummury, BioRoberta, all of which will be coupled with a CRF layer as tag encoding of named entities.

These variants are powerful in feature extraction in the medical domain, providing advantages for clinical notes especially as we deal with long reports which makes its feature extraction or context encoding difficult.

\begin{figure}[htbp]
\centering
  \includegraphics[width=5in]{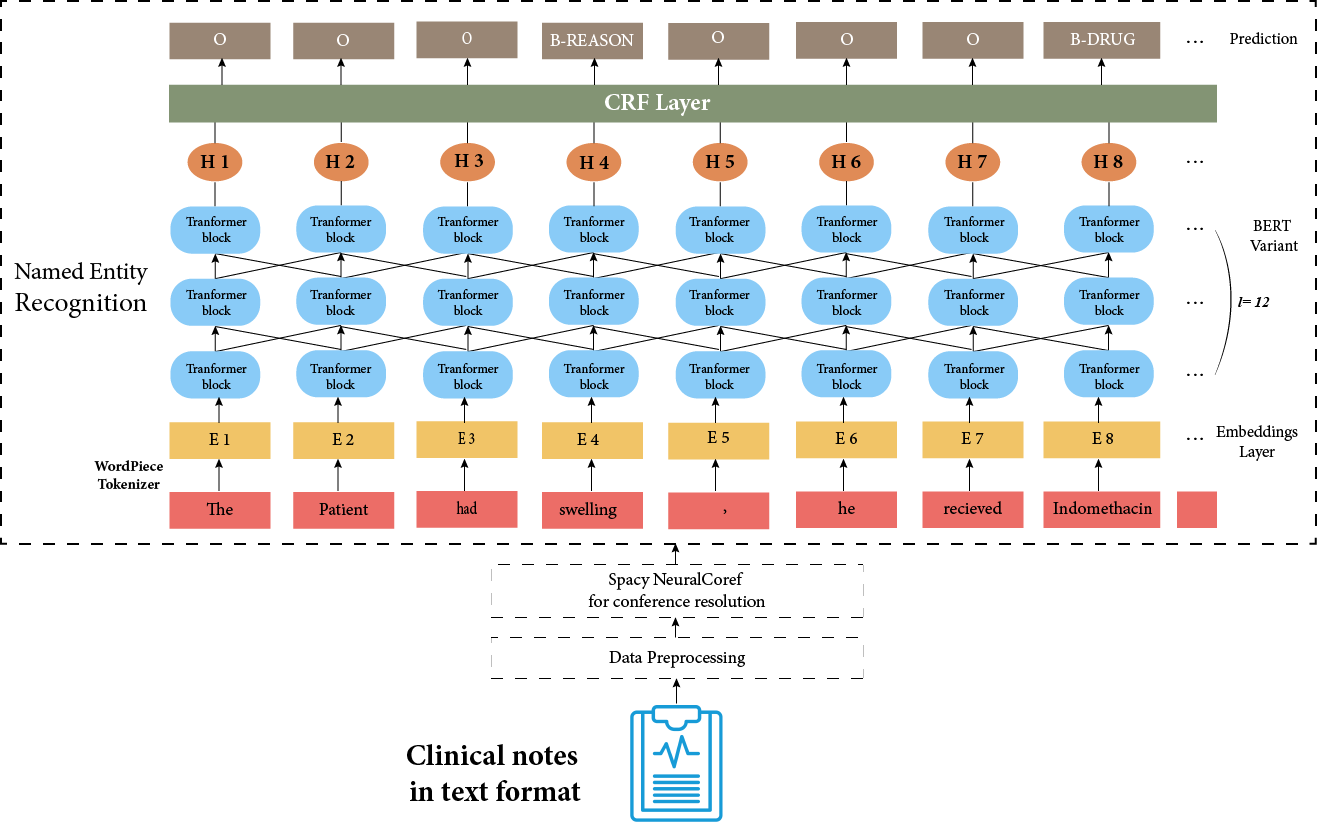}
 \caption{Named Entity Recognition Module}
  \label{fgr:ner}
\end{figure}

Thus, we adopt the following configuration, as shown in Figure \ref{fgr:ner}:
\begin{itemize}
\item 12 stacked encoders.
\item 768 as the hidden size representing the number of features of the hidden state for BERT.
\item 12 heads in the MultiHead attention layers.
\item 128 is the size of the BERT neural network input data.
\item 17 is the batch size representing the number of samples that will be propagated through the network.
\item 10 for the number of epochs which defines the number of times the learning algorithm will train on the training data set.
\item A CRF layer which allows to link the result obtained with the appropriate class of the named entity.
\end{itemize}

\subsubsection{Co-reference Resolution}

For coreference resolution, we apply the NeuralCoref model which is a pipeline extension for spaCy 2.1+ that annotates and resolves coreference groups using a neural network.
\subsubsection{Relation Extraction}

We choose to use BERT variants in this task because the learning process is much faster than that of the LSTM or CNN models since we deal with pre-trained models that only require adjustments. Moreover, the knowledge of the biomedical domain in these variants of the models is an advantage compared to other methods.

Relationship extraction (RE) is a binary classification problem, unlike the named entity recognition task, where it was used for token classification, we will test BERT variants that use the concept of sequence classification to predict relationships.

\begin{figure}[htbp]
\centering
  \includegraphics[width=5in]{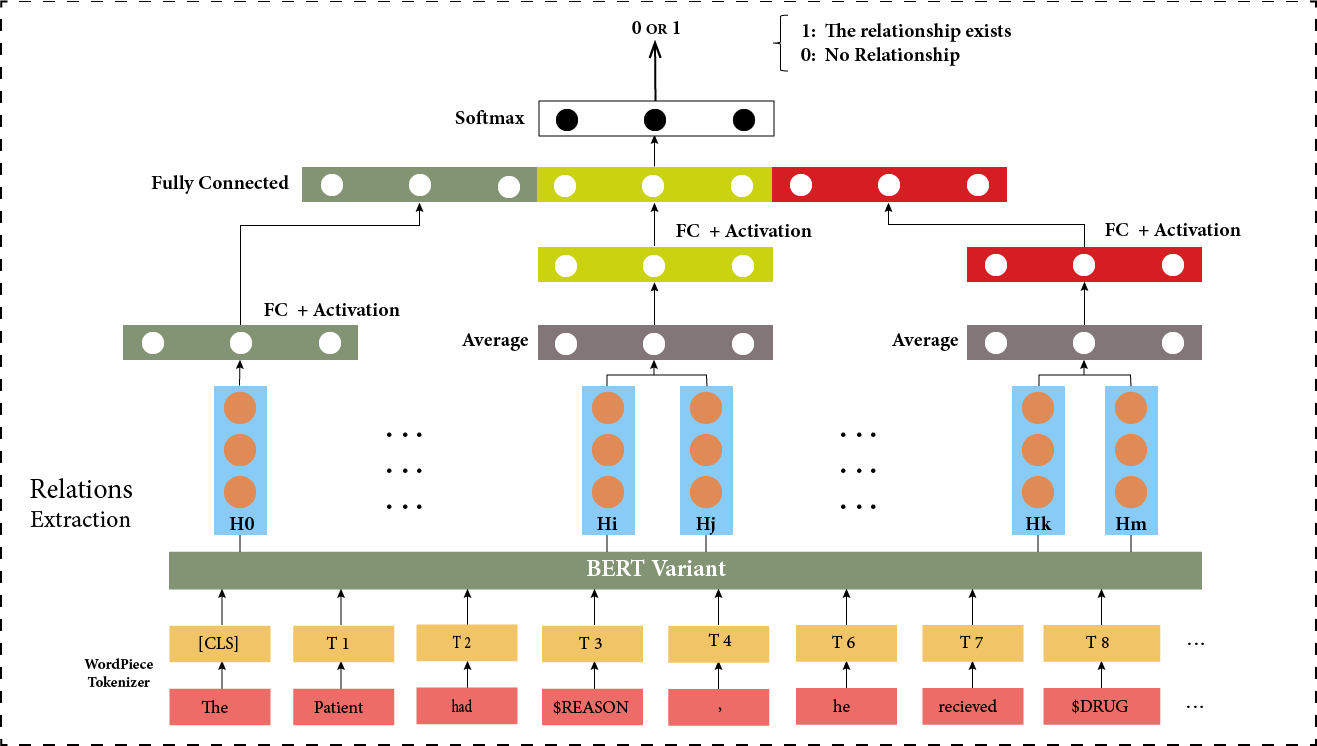}
 \caption{Relationship Extraction Module}
  \label{fgr:re}
\end{figure}

In sequence classification, a representation of the input sentence, called the CLS token (for classification) is obtained. This CLS token, which essentially contains information about the words and context of the entire sentence, is fed into a fully connected neural network that implements the binary classification task.

\subsection{Graph Construction}

In this section, we present the conceptual design of the data model for the knowledge graph. The model is based on the existing labels of the entities in the dataset, which we have organized and structured in consultation with healthcare professionals.

The objective of this model is to analyze the interactions of drugs and their side effects. The model models 5 types of entities with their attributes that will be analyzed in a concise and precise way.

The data model is schematized as shown in the following figure \ref{fgr:datamodel}:

\begin{figure}[htbp]
\centering
  \includegraphics[width=4.5in]{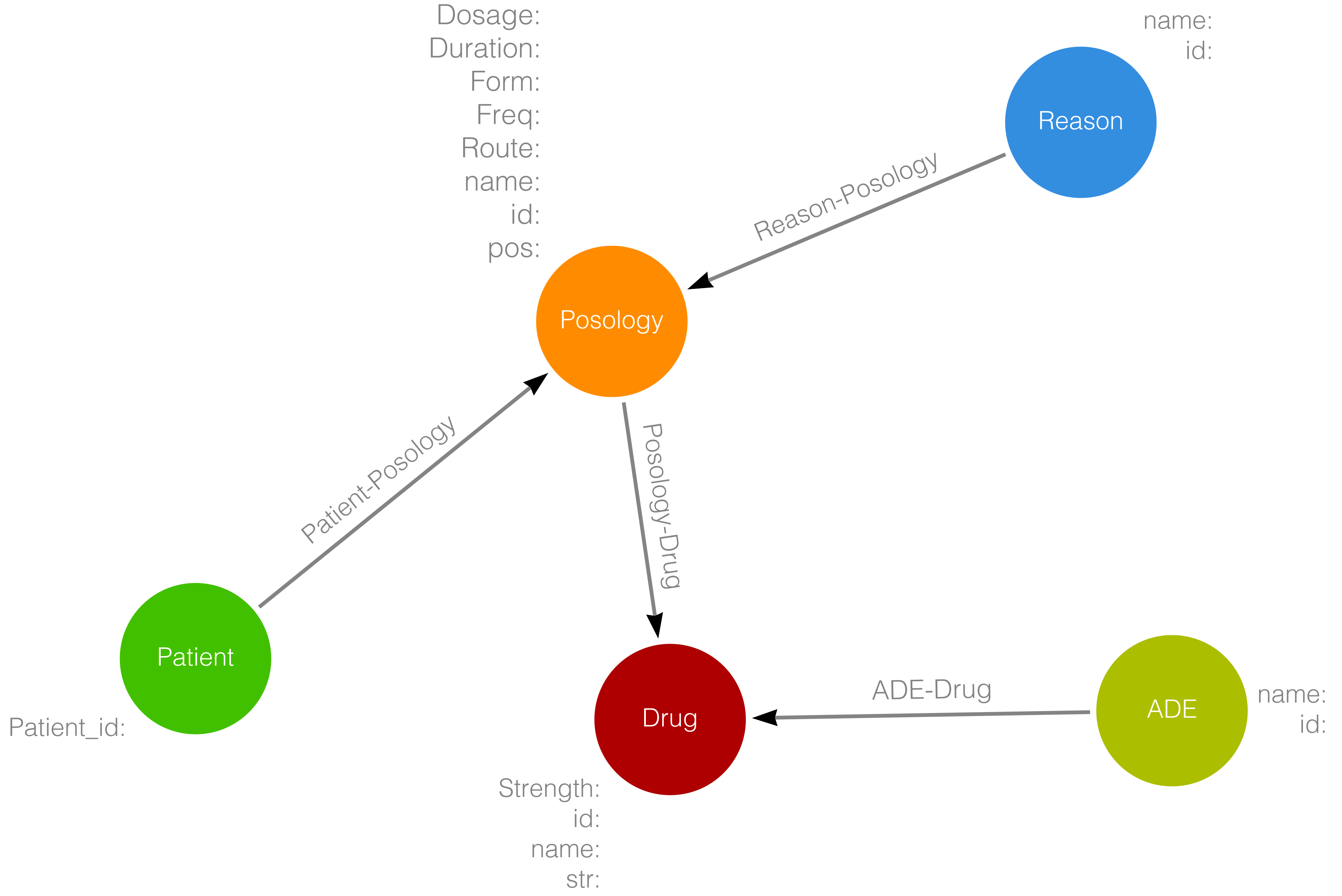}
 \caption{Graph Data Model}
  \label{fgr:datamodel}
\end{figure}

In this conceptual data model:
\begin{itemize}
\item The node "\textbf{POSOLOGY}" represents the prescription of the drug, it contains 8 attributes namely: dosage, duration, form, frequency, route of administration of the drug (route), name and identifier (id) of this node and special attribute (pos) which is an expression that gathers all the attributes of the node in a single string.
\item The node "\textbf{PATIENT}" which contains a single attribute which is the identifier of this patient.
\item The node "\textbf{DRUG}" represents the drug, it contains three attributes namely: the drug concentration (Strength), the node identifier, and the drug name and the special attribute (str) which collects all the attributes of the node into a single string.
\item The node "\textbf{ADE}" which represents the side effect, it contains two attributes namely: the name of the side effect and the node identifier.
\item The node "\textbf{REASON}" which represents the reason for taking this drug, it contains two attributes namely: the name of the reason and the node identifier.
\end{itemize}

\begin{table*}[htbp]
\footnotesize
\caption{Graph Data Model Description}
\label{tab:gdm}
\begin{tabular}{|p{1cm}|p{3.5cm}|p{1.5cm}|p{8cm}|}
\toprule
\textbf{Node}             & \textbf{Node Description}                                                                                   & \textbf{Attribute} & \textbf{Attribute Description}                                                    \\ \midrule
\multirow{8}{*}{Posology} & \multirow{8}{*}{\begin{tabular}[c]{@{}l@{}}The prescription \\ of the drug\end{tabular}}                                                               & Dosage             & The amount of a specific dosage of medication taken at one   time                 \\ \cmidrule(l){3-4} 
                          &                                                                                                             & Duration           & The period of time for which the drug is effective                                \\ \cmidrule(l){3-4} 
                          &                                                                                                             & Form               & The form of the drug in which it is marketed for use                              \\ \cmidrule(l){3-4} 
                          &                                                                                                             & Freq               & The frequency at which doses of the drug are administered                         \\ \cmidrule(l){3-4} 
                          &                                                                                                             & Route              & The route of administration of the drug                                           \\ \cmidrule(l){3-4} 
                          &                                                                                                             & name               & The drug name                                                                     \\ \cmidrule(l){3-4} 
                          &                                                                                                             & id                 & A unique number that identifies the node                                          \\ \cmidrule(l){3-4} 
                          &                                                                                                             & pos                & An expression that combines all the attributes into a single string \\ \midrule
\multirow{4}{*}{Drug}     & \multirow{4}{*}{\begin{tabular}[c]{@{}l@{}}The information \\ of the drug\end{tabular}}                                                                & Strength            & The drug concentration                                                            \\ \cmidrule(l){3-4} 
                          &                                                                                                             & name               & The drug name                                                                     \\ \cmidrule(l){3-4} 
                          &                                                                                                             & id                 & A unique number that identifies the node                                          \\ \cmidrule(l){3-4} 
                          &                                                                                                             & str            & An expression that combines all the attributes into a single string \\ \midrule
Patient                   & The patient concerned by the clinical note                                                    & id                 & A unique number that identifies the node                                          \\ \midrule
\multirow{2}{*}{Reason}   & \multirow{2}{*}{\begin{tabular}[c]{@{}l@{}}The reason for the\\drug intake\end{tabular}}                                                             & name               & The name of the reason                                                            \\ \cmidrule(l){3-4} 
                          &                                                                                                             & id                 & A unique number that identifies the node                                          \\ \midrule
\multirow{2}{*}{ADE}      & \multirow{2}{*}{\begin{tabular}[c]{@{}l@{}}A manifestation of a side\\ effect of the drug\end{tabular}} & name               & The name of the side effect                                                       \\ \cmidrule(l){3-4} 
                          &                                                                                                             & id                 & A unique number that identifies the node                                          \\ \bottomrule
\end{tabular}
\end{table*}

\subsection{Graph Analysis}

The main objective of this paper is to propose a general approach for the construction of a biomedical knowledge graph that can be used as a basis for analysis. In order to illustrate the interest of this construction we have applied a set of algorithms for its analysis.

\section{Results and Evaluation}

\subsection{Dataset Description}

The data used in this study is openly accessible and consists of 505 patient clinical notes from the Medical Information Mart for Intensive Care-III (MIMIC-III) clinical care database \cite{johnson2016mimic}. These records were selected using a query that searched for a clinical note in the International Classification of Diseases code description of each record. Each record in the dataset was annotated by two independent annotators and a third annotator resolved conflicts.

\subsection{Tools and Environment}

For the implementation of the solution, we used neo4j for the construction of the graph database and the Cypher query language for its analysis \cite{guia2017graph}. Cypher is a declarative query language for graphs that uses graph pattern matching as the primary mechanism for selecting graph data (for read-only and mutation operations).

The environment used for the training and execution of the models is a Linux Ubuntu version 20.04, Core i7, 16G DDR4, dual graphics cards AMD RADEON 4G and INTEL 1G.

For the execution time, and due to the efficiency of neo4j, the generation of the graphs required about 5 to 10 seconds, and the analysis query on the generated graphs about 1 to 2 seconds, depending on the subgraphs used and the size / complexity of the corresponding texts.

\subsection{Validation and Results}

\subsubsection{Configuration}

As with the Named Entity Recognition model, in order to train and test the relationship extraction models, the data had to be transformed into a particular format.

After dividing the training data into train and dev, each record was divided into paragraphs using the same method used for NER. The next step was to map each drug entity to all other possible entities in that paragraph. This produces a list of all possible relationships in that paragraph. Once the list was obtained, each entity text was replaced with @entity-typ\$. For example, the drug 'Lisinopril' is replaced by @Drug\$ and '20mg' by @Strength\$. This was performed for each relationship, which means that each data point has only one relationship, i.e. only one pair of entities.

Finally, a label indicating whether the entities in this text are related has been added: 1 representing a relationship and 0 otherwise.

\subsubsection{Results}

We apply this configuration for the 4 BERT variants and compare their performances.

\begin{table}[htbp]
\centering
\small
\caption{Models Performances}
\label{tab:xx}
\begin{tabular}{@{}lll@{}}
\toprule
\textbf{Task}        & \textbf{Deep Learning Model} & \textbf{F1 Score} \\ \midrule
\multirow{4}{*}{NER} & BioBERT+CRF                  & 89\%              \\ \cmidrule(l){2-3} 
                     & Bio\_ClinicalBERT+CRF        & \textbf{90.7\%}   \\ \cmidrule(l){2-3} 
                     & Bio\_Discharge\_Summary+CRF  & 87\%              \\ \cmidrule(l){2-3} 
                     & Biomed\_roberta\_base+CRF    & 84\%              \\ \midrule
\multirow{4}{*}{RE}  & BioBERT                      & 85\%              \\ \cmidrule(l){2-3} 
                     & Bio\_ClinicalBERT            & \textbf{88\%}     \\ \cmidrule(l){2-3} 
                     & Bio\_Discharge\_Summary      & 81\%              \\ \cmidrule(l){2-3} 
                     & Biomed\_roberta\_base        & 79\%              \\ \bottomrule
\end{tabular}
\end{table}

\subsection{Graph Construction}

Then, we build a graph database on neo4j that we analyze using the Cypher query language and the analysis algorithms for graphs.

We illustrate here a sub-graph built from 17 clinical notes, i.e. the information from 17 patients (Fig \ref{fgr:subg}). The database thus includes 4552 nodes and 14121 relations as shown in the following figure:

\begin{figure}[h]
\centering
  \includegraphics[width=4in]{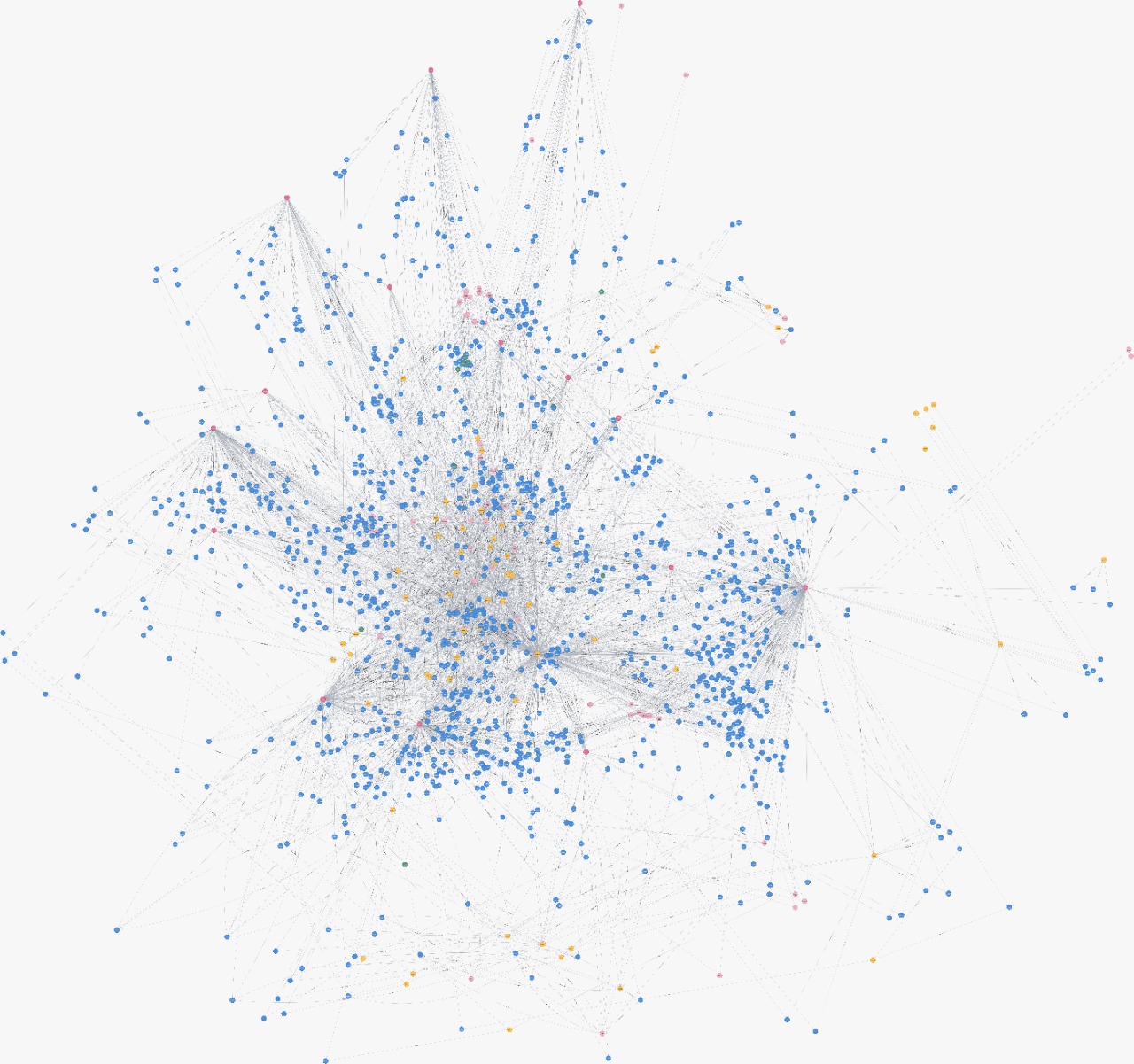}
  \caption{Sub-Graph Generated of 17 clinical notes}
  \label{fgr:subg}
\end{figure}

\begin{figure}[htbp]
\centering
  \includegraphics[width=6in]{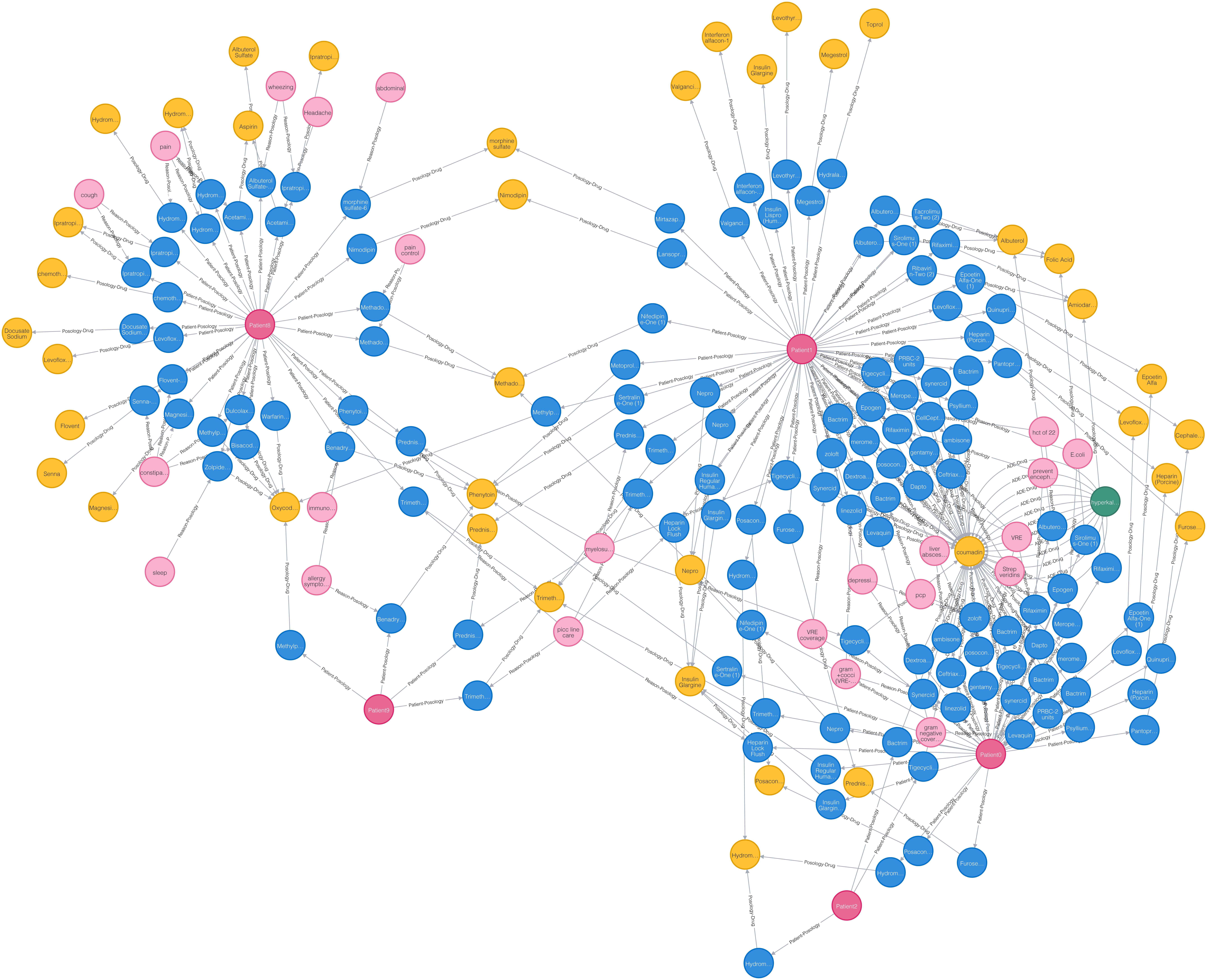}
  \caption{Sub-Graph Excerpt}
  \label{fgr:subge}
\end{figure}

\subsection{Graph Analysis}

An end-to-end pipeline was created to transform raw clinical notes into a structured and more intuitive form. First, the raw clinical note is preprocessed for coreference resolution task. This preprocessed data is then fed to the BioClinicalBERT+CRF model for NER to obtain predictions about the entities present in the clinical note. Using these predicted entities, the raw clinical note is again preprocessed for the relationship extraction model which is BioClinicalBERT+Softmax.

After obtaining the predictions of the relationships between the entities, a table is generated to associate each drug with all the entities related to it.

From the constructed graph, it is possible to analyze, among others: 
\begin{itemize}
\item the drug most used by the patients, i.e. the drug node (DRUG) which has the maximum number of incident links.
\item the most important reason for taking the drug in the graph, to do this we will apply the same algorithm of degree centrality but this time for the reason type nodes (REASON).
\item the set of prescriptions that treat the same symptom. That is, the sets of prescription and reason type nodes that can reach any other node by crossing edges.
\item if there are several side effects from the same drug, for this we will detect the compound communities on the "ADE-Drug" relationships.
\end{itemize}

Hereby, in this extract, it appears that the drug "coumadin", which is an organic compound of the coumarin family, represents the most important drug in the graph because it has a large number of edges that connect to it.

Then, the node "constipation" which represents the fact when one has a bowel movement less than three times a week. And that can be caused by a diet not rich enough in fiber, lack of exercise or pathologies such as hypothyroidism is the most important reason type node.

Finally, it turns out that 7 different side effects are from the same drug which is "coumadin". So this drug must be closely monitored.

\begin{figure}[htbp]
\centering
  \includegraphics[width=4in]{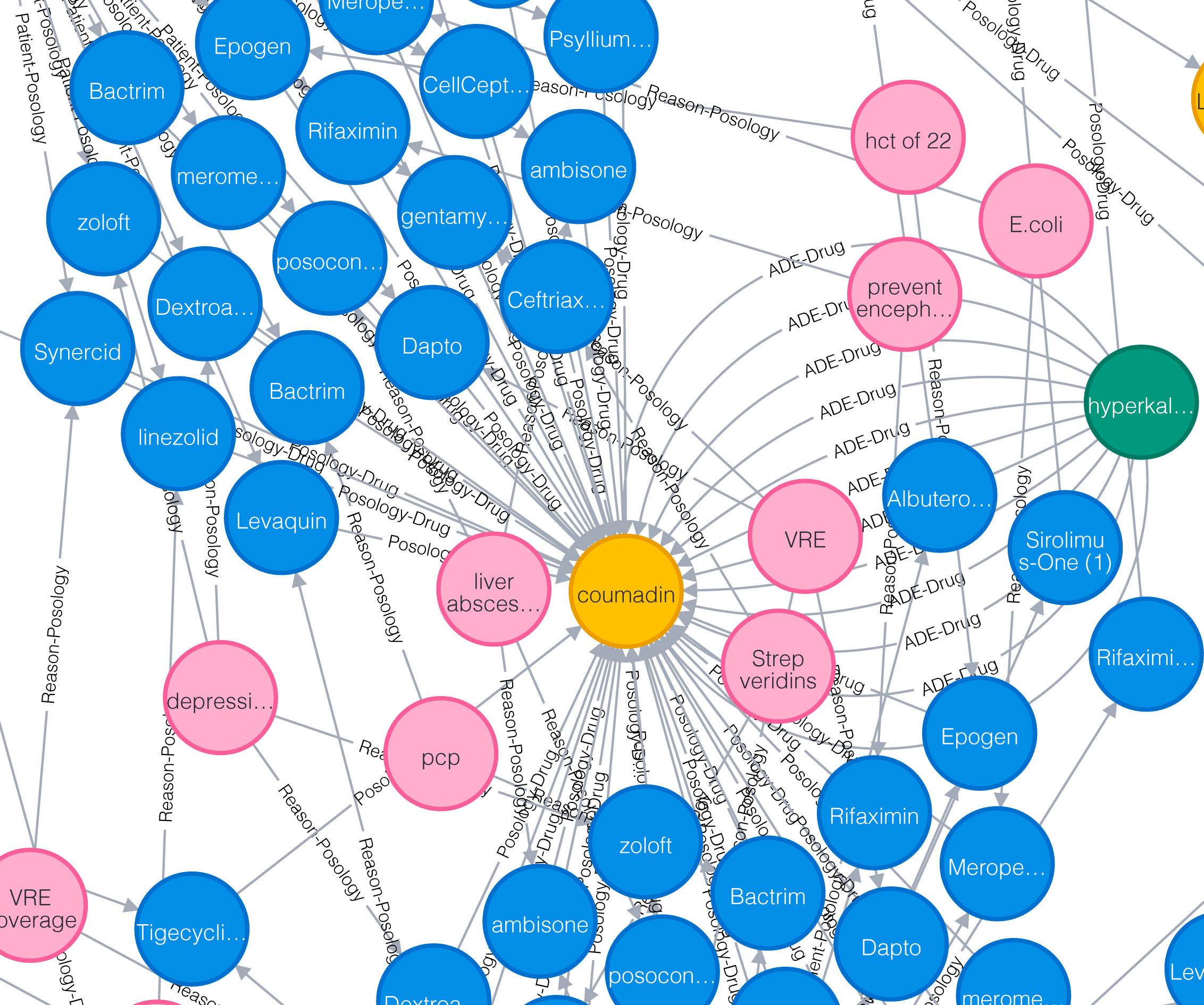}
  \caption{Zoom on Sub-Graph Excerpt}
  \label{fgr:coum}
\end{figure}

In figure \ref{fgr:coum}, we zoomed on the corresponding area of the graph. This part of the analysis is variable depending on the chosen extract.

\section{Conclusion}

In this paper we propose a complete and accurate approach for the construction of a biomedical knowledge graph from any clinical data providing an exploitable basis for analysis.

To recall that the challenge raised by the exploitation of medical textual data is that the information on patients is massive, some of the tasks carried out by health professionals are repetitive, long and tedious like the search for information. This need has led to the search for other more innovative and faster solutions such as knowledge graph-based information retrieval. 
Most of the works in the literature focus on a specific part in the life cycle of building a knowledge graph, however they do not combine these elements for the global construction of this graph and sometimes they do not start the analysis part to illustrate the interest of the construction.

As a perspective, we aim to develop a search engine for doctors and patients in order to benefit from the functionalities of this knowledge graph for the contextual search of the information they need. On the other hand, we hope to widen the application domain of this knowledge graph in order to be used in other domains, as the pharmaceutical domain. 

\bibliographystyle{unsrt}  
\bibliography{template}  

\end{document}